\documentclass{article}
\usepackage{ijcai23}

\usepackage{times}
\usepackage{soul}
\usepackage{url}
\usepackage[hidelinks]{hyperref}
\usepackage[utf8]{inputenc}
\usepackage[small]{caption}
\usepackage{graphicx}
\usepackage{amsmath}
\usepackage{amsthm}
\usepackage{booktabs}
\usepackage{algorithm}
\usepackage{algorithmic}
\usepackage[switch]{lineno}

\usepackage{amssymb}
\usepackage{threeparttable}
\usepackage{epstopdf}
\usepackage{stfloats}
\usepackage{subfigure}
\usepackage{multirow}
\usepackage{makecell}
\usepackage{pifont}
\usepackage{dsfont}

\title{Cluster-aware Contrastive Learning for Unsupervised Out-of-distribution Detection}

\author{
Menglong Chen$^1$
\and
Xingtai Gui$^1$
\and
Shicai Fan$^{1,2}$
\affiliations
$^1$University of Electronic Science and Technology of China(UESTC)\\
$^2$Shenzhen Institute of Advanced Study, UESTC\\
\emails
\{menglongchen, tabgui\}@std.uestc.edu.cn,
shicaifan@uestc.edu.cn,
}

\begin{document}

\maketitle

% no more 200words
\begin{abstract}
    \textit{Unsupervised out-of-distribution (OOD) Detection} aims to separate the samples falling outside the distribution of training data without label information. Among numerous branches, contrastive learning has shown its excellent capability of learning discriminative representation in OOD detection. However, for its limited vision, merely focusing on instance-level relationship between augmented samples, it lacks attention to the relationship between samples with same semantics. Based on the classic contrastive learning, we propose \textit{Cluster-aware Contrastive Learning (CCL)} framework for unsupervised OOD detection, which considers both instance-level and semantic-level information. Specifically, we study a cooperation strategy of clustering and contrastive learning to effectively extract the latent semantics and design a cluster-aware contrastive loss function to enhance OOD discriminative ability. The loss function can simultaneously pay attention to the global and local relationships by treating both the cluster centers and the samples belonging to the same cluster as positive samples. We conducted sufficient experiments to verify the effectiveness of our framework and the model achieves significant improvement on various image benchmarks.
    
\end{abstract}
% -----------------------------------------------

\section{Introduction}
\label{sec:intro}

Different from the traditional closed world assumption, OOD detection requires the model to distinguish samples with different distributions from in-distribution (ID) samples. It has demonstrated huge potential in diverse fields such as autonomous driving \cite{autonomous-chan2021entropy}, industrial defect detection \cite{industrial-bergmann2019mvtec}, medical diagnosis \cite{medical-schlegl2017unsupervised}, etc.

Recent methods of OOD detection include density-based \cite{dense2-ren2019likelihood}, reconstruction-based \cite{recons2-perera2019ocgan}, and classification-based \cite{class3-hsu2020generalized} approaches, which mostly belong to supervised or semi-supervised OOD detection methods. However, further attention need be paid to the more difficult unsupervised OOD detection when the label information is completely unavailable. In unsupervised OOD detection, self-supervised methods \cite{self1-hendrycks2019using,self2-golan2018deep} learn discriminative representation by designing different auxiliary tasks and as one of them, contrastive learning method \cite{msc-reiss2021mean} has made great contributions.

The original goal of contrastive learning is to obtain individual, task-agnostic representation for each sample. It is obvious that merely caring about instance-level features does not take advantage of latent semantic information. More intuitively, it's beneficial to distinguish the OOD samples during online testing by clustering ID samples with the same semantics in a denser area. However, there is a mutual repulsion between feature extraction at the instance-level and class relationship preservation at the semantic-level in contrastive learning. How to reasonably introduce clustering to contrastive learning needs further research. Though some deep clustering methods \cite{21-li2021contrastive,17-zhang2021supporting,19-sharma2020clustering} have discussed the possibility of combining contrastive learning and clustering, these works increase the inter-class variance of training samples to ensure the classification accuracy, which are not suitable for OOD detection. And some clustering contrastive learning methods \cite{pcl,swav} do not fully extract semantic information for ignoring the local relationship between samples. Given above introduction, we further discuss the possibility of combining contrastive learning and clustering in unsupervised OOD detection.

In this paper, we propose \textit{cluster-aware contrastive learning (CCL)} framework, a novel framework not only retains discriminative representation ability for each individual instance but also has a wider vision that pays attention to the global and local similarity relationship between samples. On the one hand, it is vital to study a proper strategy to introduce clustering operation to contrastive learning. We perform clustering at appropriate location to extract semantic information effectively. And a warm-up stage acquiring enough instance representation ability is required before periodically updating the cluster centers. On the other hand, we design a cluster-aware contrastive loss function containing two parts, \textit{Cluster Center Loss} and \textit{Cluster Instance Loss}. The former one makes the augmented samples close to their corresponding cluster center while pushes them away from other cluster centers for the global relationship. And the latter one increases the similarity between samples in the same cluster for the local relationship.

The contributions of this paper can be summarized as follows:
\begin{itemize}

    \item We propose a novel unsupervised OOD detection framework, \textit{cluster-aware contrastive learning, CCL}, which makes the contrastive learning model take advantage of instance- and semantic-level information. And we discuss the clustering strategy introducing clustering to contrastive learning.

    \item We design a cluster-aware contrastive loss function. It pays attention to the global and local sample relationship by making individual samples closer to their corresponding cluster centers and the other samples within the same cluster.

    \item  We conduct plenty of comprehensive experiments and compare CCL with state-of-the-art methods. Both the comparison results and the ablation experimental results prove the effectiveness of our approach.
\end{itemize}

%---------------------------------------------------------------------

\section{Related Work}

\subsection{Unsupervised Out-of-distribution Detection}

On account of the massive amount of image data and the high cost of labeling data, semi-supervised learning or unsupervised learning in OOD detection have been received great attention. For example, ODIN \cite{04-liang2017enhancing} uses a basic pre-trained model and separates the softmax score distributions between ID and OOD images by a temperature scaling adding perturbation. \cite{12-yang2021semantically} designs a unsupervised dual grouping (UDG) to solve the semantically coherent unsupervised OOD. In particular, as one of unsupervised learning methods, self-supervised method is widely used in OOD detection. For examples, \cite{11-madan2022self} presents a self-supervised masked convolutional transformer block (SSMCTB). \cite{20-zhou2022rethinking} designs a hierarchical semantic reconstruction framework, which maximizes the compression of the auto-encode latent space. SSD \cite{23-sehwag2020ssd} uses self-supervised representation learning and Mahalanobis distance score to identify OOD samples on the pre-trained softmax neural classifier. Different from some methods based on pre-trained models, we solve the completely unsupervised OOD problem which does not introduce any prior semantic label information.

\subsection{Contrastive Learning}
In self-supervised learning, contrastive learning methods have been widely used because of excellent representation ability via instance discrimination, such as SimCLR \cite{simclr} and MoCo \cite{moco}. Contrastive learning considers two augmented views of the same instance as the positive to be pulled closer, and the rest samples are considered as the negative to be pushed farther apart. In OOD detection, CSI \cite{csi} uses data enhancement in contrastive learning to generate OOD negative samples to improve the sensitivity to OOD samples. \cite{mcl} proposes masked contrastive learning for OOD detection, performing a designed auxiliary task class-conditional mask in contrastive learning. To introduce semantic information, some researches try to add clustering as an auxiliary task to contrastive learning, such as PCL \cite{pcl} and SWAV \cite{swav}. However, it is difficult to integrate instance and semantic representation ability at the same time and there are few studies on combining clustering and contrastive learning to solve OOD problems. We design a novel framework to introduce clustering, which has a wider vision to extract semantic information from sample relationship.

\subsection{Contrastive Cluster}
Contrastive clustering is essentially a deep clustering method based on contrastive learning and has been widely used in images \cite{19-sharma2020clustering}, contexts \cite{17-zhang2021supporting} and graphs \cite{22-wang2022clusterscl}. For example, \cite{21-li2021contrastive} proposes a basic framework of contrastive clustering. The instance- and semantic-level contrastive learning are respectively conducted in the row and column feature space by a similarity loss function. And \textit{Graph Contrastive Clustering (GCC)} \cite{16-zhong2021graph} proposes a novel graph-based contrastive learning strategy to learn more compact clustering assignments. The core of contrastive clustering is performing the instance- and semantic-level contrastive learning to minimize the intra-cluster variance and maximize the inter-cluster variance, which gives a lot of inspiration for OOD detection. In OOD detection, \cite{10-albert2022embedding} applies an outlier sensitive clustering at the semantic level to detect the OOD clusters and ID noisy outliers. \cite{09-lehmann2021layer} performs a layer-wise cluster analysis at inference to identify OOD samples. Since contrastive clustering will increase the inter-class variance for classification, it is inappropriate for direct application in OOD task. However, we could refer to its idea of combining clustering and contrast learning in OOD detection.
%-----------------------------------------------------

\section{Methodology}
\label{sec:method}

% 整体框架图
\begin{figure*}[ht]
\centering
\includegraphics[width=1\linewidth]{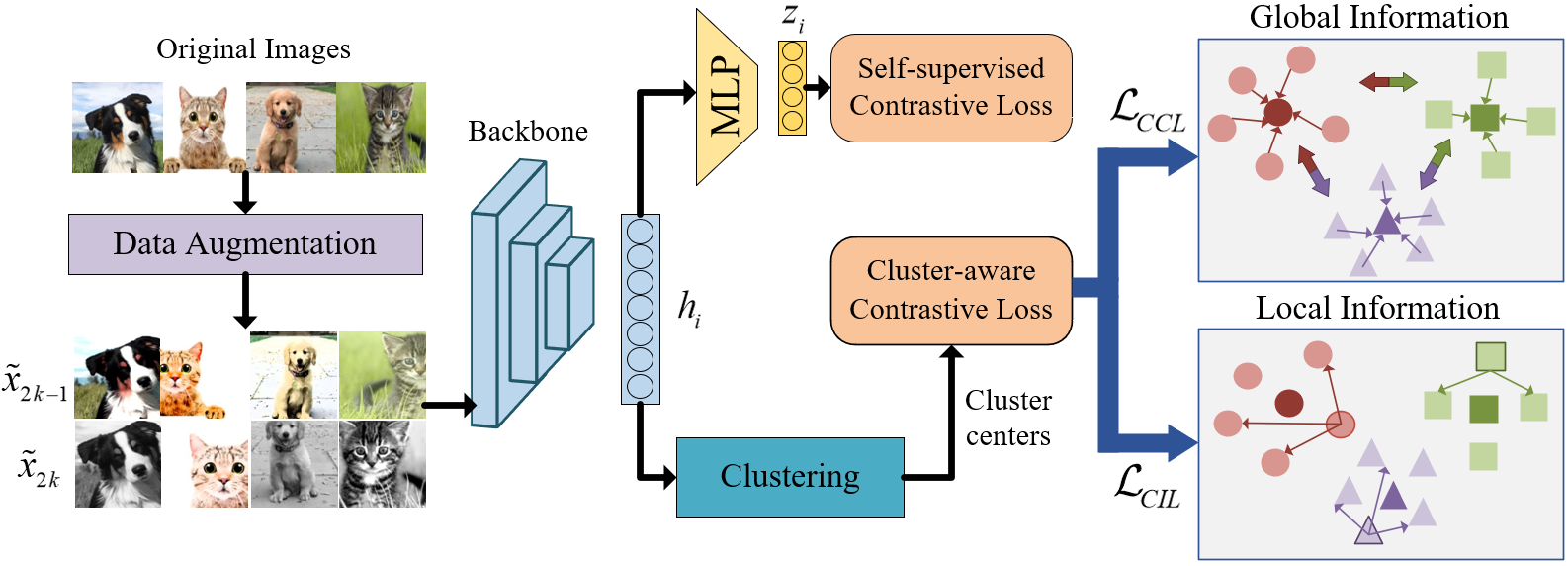}
% 说明embedding layer 和 projection layer
\caption{The framework of cluster-aware contrastive learning. The backbone encoder output $h_i$ is obtained on the last layer of encoder, called the embedding layer. In the basic contrastive learning module, the model calculates a classic self-supervised contrastive loss for instance-level representation on the last layer of MLP, called the projection layer. In the clustering module, according to the cluster centers though clustering on the embedding layer, the model calculates the designed cluster-aware contrastive loss including two parts, a cluster center loss $\mathcal{L}_{CCL}$ for global semantic information and a cluster instance loss $\mathcal{L}_{CIL}$ for local semantic information. In the test phase, a score function $score$ is required to detection OOD sample.
}
\label{fig:overview}
\end{figure*}

\subsection{Framework}
As shown in Figure \ref{fig:overview}, the proposed CCL framework contains two modules, a basic contrastive learning module and a clustering module. Considering a training set batch of $N$ unlabeled images $X = \{ {x_1},{x_2},...,{x_N}\}$, image pairs $({{\tilde x}_{2k-1}},{{\tilde x}_{2k}})$, are obtained by data augmentation including random cropping, random color distortions, random Gaussian blur same with SimCLR \cite{simclr}. The augmented image dataset is $\tilde X = \{ ({{\tilde x}_1},{{\tilde x}_2}),({{\tilde x}_3},{{\tilde x}_4})...,({{\tilde x}_{2N - 1}},{{\tilde x}_{2N}})\}$. For the sake of description, the last layer of backbone encoder $F( \cdot )$ is called \textbf{embedding layer} and the last layer of MLP $G( \cdot )$ is called \textbf{projection layer}. The augmented image ${\tilde x}_{i}$ is input into the backbone encoder $F( \cdot )$ to obtain high-dimensional feature vector $h_i$ on the embedding layer. And $h_i$ is input into MLP $G( \cdot )$ to obtain ${z}_i$ on the projection layer.

\begin{equation}
    {h}_i = F({\tilde x}_i), {z}_i = G({h}_i) = Z({\tilde x}_i)
\end{equation}

In the basic contrastive learning module, a classic self-supervised contrastive loss $\mathcal{L}_{self}$ is calculated on the projection layer,which focuses on the relationship between samples and their enhancement samples for instance-level representation. In the clustering module, for each $h_i$ on the embedding layer, the corresponding cluster center $c_i$, $i = 1,...,2N$, is obtained by K-means to calculate the cluster-aware contrastive loss $\mathcal{L}_{cluster}$. It includes two parts: a cluster center loss $\mathcal{L}_{CCL}$ and a cluster instance loss $\mathcal{L}_{CIL}$. The self-supervised contrastive loss updates the MLP and are simultaneously combined with the cluster-aware contrastive loss to update the backbone encoder. Finally, OOD detection needs a score function $score$ to identify whether a new input image is OOD sample according to its score.

\subsection{Loss Function}
\label{met:loss}
The loss function plays a decisive role in contrastive learning. Although traditional contrastive learning has strong individual discrimination capacity, it ignores the relationship among instances with similar semantic. On the contrary, Supervised contrastive learning (SupCon) \cite{supcon}, uses label information to extract the relationship between samples. Moreover, it is a natural and common way to introduce semantic information in unsupervised tasks utilizing cluster centers obtained by clustering as pseudo labels. Combining the above two advantages of using semantic information, we design a novel loss function as an auxiliary task in contrastive learning.

\subsubsection{Contrastive Loss}
The instance-level discrimination capacity of contrastive learning is obtained by self-supervised contrastive loss. We adopt the classic contrastive learning loss function in SimCLR, \cite{simclr}. For a positive pair of examples $(i,j)$, it defined as follows:

\begin{equation}
    \ell(i, j)=-\log \frac{\exp \left(s({z_i}, {z_j}) / \tau\right)}{\sum_{k=1}^{2 N} \mathds{1}_{[k \neq i]} \exp \left(s({z_i}, {z_k}) / \tau\right)}
    \label{NT-Xent}
\end{equation}
where $\mathds{1}_{[k \neq i]} \in\{0,1\}$ is an indicator function evaluating to 1 if $ k \neq i$ and $\tau$ denotes a temperature parameter. And, ${s(z_i,z_j)} = {{z}_i^{\top}{z}_j}/{\left(\left\|{z}_i\right\|\left\|{z}_j\right\|\right)}$ is the cosine similarity between the projection layer features $(z_i, z_j)$. For every sorted image pair $({{\tilde x}_{2k-1}},{\tilde x}_{2k})$, the whole self-supervised contrastive learning loss is:

\begin{equation}
  \mathcal{L}_{self}=\frac{1}{2N} \sum_{k=1}^N[\ell(2k-1,2k)+\ell(2k, 2k-1)]
  \label{equation: simclr-loss}
\end{equation}

\subsubsection{Cluster-aware Contrastive Loss}
To make the backbone encoder pay attention to the latent semantic information, we cluster the output of the encoder to obtain cluster centers as the semantic labels. The cluster-aware contrastive loss includes two parts, \textit{Cluster Center Loss} for global semantic information and \textit{Cluster Instance Loss} for local semantic information.

\paragraph{Cluster Center Loss.} Clustering is performed on the embedding layer to obtain the cluster centers. The cluster center loss takes the cluster center that a specific sample belonging to as the positive and the other cluster centers as the negative. 

Specifically, for each output of the encoder $h_i$, there is a corresponding cluster center $c_i$ calculated by K-means using all ID samples. The cluster center $c_i$ is the positive sample and the other cluster centers are the negative samples. Similar to instance-level contrastive loss function, we use the cosine similarity $s({h_i},{c_i}) = {{{h_i}^T{c_i}}}/{{||{h_i}||||{c_i}||}}$, and the cluster center loss function is defined as follows:

\begin{equation}
    \mathcal{L}_{CCL}  = \frac{1}{{2N}}\sum\limits_{i = 1}^{2N} {( - \log \frac{{\exp (s({h_i},{c_i})/{\phi _i})}}{{\sum\limits_{j = 1}^R \mathds{1}_{[c_j \neq c_i]} {\exp (s({h_i},{c_j})/{\phi _j})} }})}
\end{equation}
where $\mathds{1}_{[c_j \neq c_i]}$ means removing the cluster center to which the sample encoding $h_i$ belongs. $R$ denotes the number of cluster centers and ${\phi}$ is a temperature parameter. We adopt the same cluster concentration estimation \cite{pcl} to adaptively adjust the temperature parameter ${\phi}$ :

\begin{equation}
    \phi  = \frac{{\sum\limits_{t = 1}^T {||{h_t} - c|{|_2}} }}{{T\log (T + \alpha )}}
\end{equation}
where $T$ is the number of samples in the same cluster $c$ and ${\alpha}$ is a smooth parameter to ensure that small clusters do not have an overlarge ${\phi}$.

\paragraph{Cluster Instance Loss.} The cluster centers are the semantic information concentration of all samples and the cluster center loss focuses on the relationship between samples and cluster centers, which ignores the relationship between samples in a cluster to a certain extent. To make the model focus on the local sample relationship, we design the cluster instance loss based on SupCon \cite{supcon}. In the batch $N$, cluster instance loss takes the samples in the same cluster as the positive, and the rest samples as the negative, which can be formulated as follows:

\begin{equation}
    {\mathcal{L}_{CIL}} = \frac{1}{{|N|}}\sum\limits_{i \in N} {\frac{{ - 1}}{{|P(i)|}}\sum\limits_{p \in P(i)} {\log \frac{{\exp (s({h_i}, {h_p}) / \tau )}}{{\sum\limits_{a \in A(i)} {\exp (s({h_i}, {h_a})/\tau )} }}} }
\end{equation}
where ${\rm{A(i)}} \equiv N \backslash \{ i\}$ is the rest samples except ${\tilde x}_i$, $P(i) \equiv \{ p \in A(i):{c_i} = {c_p}\}$ is the set belonging to the same cluster center of ${\tilde x}_i$ in batch $N$. $|P(i)|$ is the set number. $\tau$ is a temperature parameter. Overall, the whole cluster-aware contrastive loss ${\mathcal{L}_{cluster}}$ takes the mean of cluster center loss and cluster instance loss, formulated as follows:

\begin{equation}
    {\mathcal{L}_{cluster}} = ({\mathcal{L}_{CCL}} + {\mathcal{L}_{CIL}})/2
\end{equation}

Finally, the loss function in training phase $\mathcal{L}$ is the combination of self-supervised contrastive loss $\mathcal{L}_{self}$ and cluster-aware contrastive loss ${\mathcal{L}_{cluster}}$ by a weight $\lambda$, which is:

\begin{equation}
    \mathcal{L} = (1 - \lambda ){\mathcal{L}_{self}} + \lambda {\mathcal{L}_{cluster}}
\end{equation}

%-----------------------------------------------------------

\subsection{How clustering works}
\label{met:cluster}

Though the idea that combining contrastive learning and clustering is not proposed for the first time, how to make it appropriate for OOD detection needs further discussion. Recent research \cite{game} concludes that the data clustering performance of the self-supervised contrastive learning is not as good as supervised contrastive learning, which means it is potential to improve clustering ability without label information. Given that most works appreciate the output feature of embedding layer rather than projection layer, we first propose an assumption that the performance of clustering on the embedding layer is better than that on the projection layer. Further, to guarantee clustering more stably, we think it is necessary to commit a warm-up operation and update the cluster centers periodically in training phase. Detailed discussions of the above ideas are as follows:

\paragraph{Where to cluster} Given that most existing works \cite{simclr,game} operate the downstream task fine-tune or linear evaluation on the embedding layer. It is reasonable to believe that the high-dimensional features on the embedding layer contain more semantic information than the low-dimensional features on the projection layer. 

Moreover, considering reducing the distance between a sample and its augmented sample will increase the distance between the sample and other samples with same semantics, we think the model optimizations of self-supervised contrastive loss and cluster-aware contrastive loss are mutually exclusive to some extent. Specially, the former one only focuses on pulling different data augmentation versions of a sample closer while the samples with similar semantics may be pushed away and the latter one focuses on bringing the samples within a same cluster closer. Based on the above analysis, when basic contrastive learning module is on the projection layer, clustering module on the embedding layer can effectively agglomerate semantic information and make model optimization possible.

\paragraph{When to cluster} In CCL framework, we treat clustering as an auxiliary task. 
Therefore, we think only after the model obtains enough instance representation ability through the basic contrastive learning module, then it could focus on the relationship between samples by clustering, which means a warm-up stage is required at beginning.

In addition, according to the cluster centers obtained on the embedding layer, the cluster-aware contrastive loss is calculated to optimize the backbone encoder. We assume that updating cluster centers too frequently is not conducive for the encoder to extract semantics. Therefore, the updating frequency of cluster centers needs to be determined. Overall, CCL needs a warm-up at beginning and cluster centers need to be updated periodically.

Further, we'll prove our viewpoints though detailed experiments and the result of ablation experiments will be show in Section \ref{ablation}.

% ------------------------------------------------------

\subsection{Score function}
In OOD detection, the score function is finally used to identify whether a sample is an OOD sample. The commonly using Mahalanobis distance as the score function is not suitable when the number of training samples is small. The another common score function proposed in CSI \cite{csi} uses L2-norm and cosine similarity, formulated as follows:

\begin{equation}
    {score_{cos}}= \mathop {\max }\limits_m {\mathop{\rm sim}\nolimits} (Z({{\tilde x}_m}),Z({\tilde x})) \cdot ||Z({{\tilde x}_m})||
\end{equation}

Based on the score function ${score_{cos}}$, we design a novel score function by combining cosine similarity and ID sample variance, described as follows:

\begin{equation}
    {score_{{\mathop{\rm var}} }} = {score_{{\rm{cos}}}}/\sqrt {\frac{1}{{K - 1}}\sum\limits_{v \in V} {{{(Z({{\tilde x}_v}) - \bar Z({{\tilde x}_v}))}^2}} }
\end{equation}

where the denominator is the standard deviation of ID set $V$. $V \equiv \{ {{\tilde x}_v} \in \{ {{\tilde x}_m}\} ,v = 1,...,K\}$ is the first $K$ ID samples with highest score $score_{cos}$. ${\bar Z({\tilde x}_v)}$ is the sample mean.

% top{\mathop{\rm sim}\nolimits}
%-------------------------------------------------------------------------

%------------------- table: main result----------
\begin{table*}[htbp]
\centering
\begin{threeparttable}
    \setlength{\tabcolsep}{2mm}{
    \begin{tabular}{{l}*{8}{c}}
    \toprule
     &  & \multicolumn{7}{c}{CIFAR-10}                                    \\ 
    \cmidrule{3-9}
    Method & Score Function & SVHN & LSUN & ImageNet & LSUN(Fix) & ImageNet(Fix) & CIFAR-100 & Interp. \\ 
    \midrule
    Rot$\ast$   & - & 97.6 & 89.2 & 90.5 & 77.7      & 83.2          & 79.0      & 64.0    \\
    Rot+Trans$\ast$  & - & 97.8 & 92.8 & 94.2 & 81.6      & 86.7          & 82.3      & 68.1    \\
    GOAD$\ast$  & - & 96.3 & 89.3 & 91.8     & 78.8      & 83.3          & 77.2      & 59.4    \\
    SSD$\ast$  & - & 99.6 & -    & -        & -         & -             & 90.6      & -       \\
    CSI$\ast$  & $score_{cos}$ & \textbf{99.8} & 97.5 & \textbf{97.6}     & 90.3      & 93.3      & 89.2      & 79.3    \\  
    \hline
    CSI$\dagger$   & $score_{cos}$ & 99.7 & 97.3 & 97.3     & 92.8      & 95.2          & \textbf{91.4}      & 82.5    \\
    SimCLR$\dagger$  & $score_{cos}$ & 97.5 & 96.7 & 93.5     & 97.3      & 95.8    & 90.5      & 85.9    \\
    % SimCLR$\dagger$  & $score_{var}$ & 97.7 & 97.0 & 93.8     & \textbf{97.8}     & \textbf{96.0}    & 90.3      & 86.4    \\    
    CCL   & $score_{cos}$ & 98.0 & 98.4 & 96.4    &  \textbf{97.8}   & \textbf{96.0}    & 91.2      & \textbf{87.2}    \\
    CCL   & $score_{var}$ & 98.4 & \textbf{98.5} & 96.6    &  \textbf{97.8}   & \textbf{96.0}     & 91.2      & \textbf{87.2}    \\
    \toprule
    \end{tabular}}
    
    \begin{tablenotes}
    \item $\ast$  denotes results shown in CSI \cite{csi} and $\dagger$ denotes results we reproduced.
    \end{tablenotes}
\end{threeparttable}
\caption{Unsupervised OOD detection performance (AUROC \%) $\uparrow$.}
\label{tab:main}
\end{table*}
% --------------------------------------------

\section{Experiments}
\label{sec:Experiment}

We compare CCL with some competitive baselines in Section \ref{results} and perform some ablation experiments to validate our framework in Section \ref{ablation}. The detailed experimental settings are as follows:

\paragraph{Datasets and Evaluation metrics.} To compare with the SOTA methods, the dataset setting is consistent with CSI \cite{csi}. We take the unlabeled CIFAR-10 training sub-dataset as the ID training data. In the test phase, the OOD datasets include SVHN, resized LSUN and ImageNet, the fixed version of LSUN and ImageNet, CIFAR-100, and linearly-interpolated samples of CIFAR-10 (Interp.). Like many OOD researches, we report the area under the receiver operating characteristic curve (AUROC) as the threshold-free evaluation metric.

\paragraph{Implementation details.} We adopt ResNet-18 as the backbone encoder for all experiments. Specifically, the model is trained for 1000 epochs using self-supervised contrastive loss at the warm-up stage, and then for 1000 epochs using both self-supervised and cluster-aware contrastive loss function. The number of cluster centers is set to 10 and the batch size is 128. The learning rate is 0.1 with a CosineAnnealingLR scheduler and has a 50 epoch warm-up, which is same with CSI \cite{csi}. The CCL is compared with various methods including, Rot \cite{self1-hendrycks2019using}, GOAD \cite{goad-bergman2020classification}, SSD \cite{23-sehwag2020ssd}, SimCLR \cite{simclr} that is the base framework of CCL, and CSI \cite{csi} that is a competitive contrastive learning method. 

%---------------------------------------

\subsection{Main Results}
\label{results}

The main experiment results are shown in Table \ref{tab:main}. We can make the following observations from it: 

\begin{itemize}
    \item Compared with SimCLR that only considers instance-level representation, CCL outperforms on all OOD datasets, which indicates that it is beneficial to take semantic information into account for unsupervised OOD detection and our framework could take advantage of semantic information to improve representation ability of the backbone encoder.
    \item Among the methods, CCL achieves the best performance on four OOD datasets including LSUN(98.5\%), LSUN-Fix(97.8\%), ImageNet-Fix(96.0\%) and Interp(87.2\%). Specially, CCL with $score_{var}$ surpasses the CSI in \cite{csi} by 7.9\% on Interp, by 7.5\% on LSUN-Fix, by 2.7\% on ImageNet-Fix and by 2\% on CIFAR-100. On the other two datasets, SVHN and ImageNet, CCL is slightly inferior to CSI. Although CCL does not perform best on all datasets, it shows greater robustness. Compare to CSI that needs designed specific OOD augmentation approaches, CCL is more general and can perform unsupervised OOD detection on more types of datasets.
    \item From the comparison of scoring functions, it can be seen that the designed variance scoring function can improve performance to a certain extent, which is helpful for unsupervised OOD detection.
\end{itemize}

%-------------------------------------------------------------------------

%------------------- table----------
\begin{table}[tb]
\centering
\resizebox{\linewidth}{!}{
    \begin{tabular}{*{3}{l}*{4}{c}}
    \toprule
    & & & \multicolumn{4}{c}{CIFAR-10}             \\
    \cmidrule{4-7}
    $\mathcal{L}_{self}$ & $\mathcal{L}_{CCL}$ & $\mathcal{L}_{CIL}$ &SVHN & LSUN & ImageNet  & CIFAR-100 \\ \midrule
    \ding{52} &  &  & 97.7  & 97.0  & 93.8  & 90.3   \\
    \ding{52} & \ding{52} &  & 98.6  & 98.0  & 96.5  & 91.2  \\
    \ding{52} &  & \ding{52} & 98.3  & 97.7  & 95.9  & 91.5  \\
    \ding{52} & \ding{52} & \ding{52} & 98.4  & 98.5  & 97.8 & 91.2 \\
    \bottomrule
    \end{tabular}
}
\caption{Comparison of different loss functions.}
\label{tab:loss}
\end{table}
% ----------------------------------------

\subsection{Ablation Experiments}
\label{ablation}
In this section, we carry out some targeted ablation experiments to verify the effectiveness of our framework and the rationality of our assumptions in Section \ref{sec:method}, including four parts as follows:

\paragraph{Loss function.} 
To verify the effectiveness of cluster-aware contrastive loss function, we compare the performance of the model using different loss functions. The results in Table \ref{tab:loss} show the combination of self-supervised and cluster-aware contrastive loss function has the best performance on LSUN and ImageNet. Although the joint loss function $\mathcal{L}$ is 0.2\% lower than the maximum value on SVHN and 0.3\% lower on CIFAR-100, it shows greater robustness. Overall, only when self-supervised and cluster-aware contrastive loss function jointly update the backbone encoder, the model has the best performance. It is confirmed that the combination of instance and semantic recognition maximizes unsupervised OOD detection capability in our framework.
%-------------------------------------------------------------------------

%------------------- table----------
\begin{table}[tb]
\centering
\resizebox{\linewidth}{!}{
    \begin{tabular}{{l}*{4}{c}}
    \toprule
    & \multicolumn{4}{c}{CIFAR-10}   \\
    \cmidrule{2-5}
    Layer & SVHN & LSUN & ImageNet  & CIFAR-100\\
    \midrule
    SimCLR       & 97.7  & 97.0  & 93.8   & 90.3    \\    
    Projection  & 97.4  & 97.9  & 95.3   & 91.4   \\
    Embedding   & 98.4  & 98.5  & 97.8   & 91.2    \\
    \bottomrule
    \end{tabular}
}
\caption{Clustering on Embedding layer vs Projection layer.}
\label{tab:layer}
\end{table}
%-------------------------------------------------------------------------

%------------------- table----------
\begin{table}[tb]
\centering
\resizebox{\linewidth}{!}{
    \begin{tabular}{*{6}{c}}
    \toprule
    & & \multicolumn{4}{c}{CIFAR-10}             \\
    \cmidrule{3-6}
    {\small Epoch} & {\small Warm-up} & {\small SVHN} & {\small LSUN} & {\small ImageNet} & {\small CIFAR-100} \\ \midrule
    10  & \ding{56}  & 98.2  & 98.0  & 95.3 & 89.9   \\
    10  & \ding{52} & 98.4  & 98.5  & 97.8 & 91.2  \\
    0   & \ding{52} & 97.9  & 96.8  & 93.9  & 90.7   \\
    1   & \ding{52} & 97.9  & 97.8  & 96.9  & 91.3   \\
    50  & \ding{52} & 98.2  & 97.5  & 95.2  & 91.2   \\
    \bottomrule
    \end{tabular}
}
\caption{Update method of cluster centers. The warm-up stage is training the model using self-supervised contrastive loss function for 1000 epochs in advance. The Epoch is the epoch interval of updating cluster centers.}
\label{tab:update}
\end{table}
%-------------------------------------------------------------------------

% ----------------------figure--------------------------------
% 加一个PCA降维图
\begin{figure} [tb]
\centering
\subfigbottomskip=-4pt
\subfigcapskip=-3pt
    \subfigure[]{
        \includegraphics[scale=0.29]{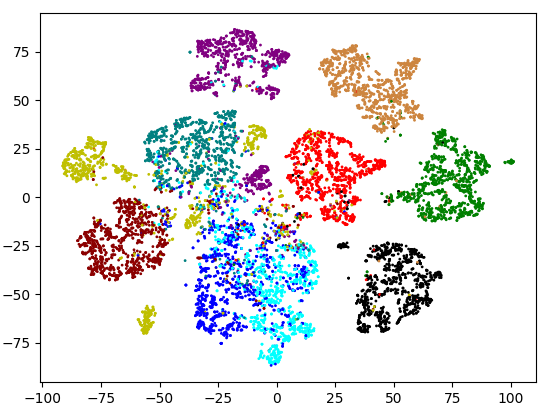}
        \label{fig:a}
    }\hspace{-8pt}
    \subfigure[]{
    \includegraphics[scale=0.29]{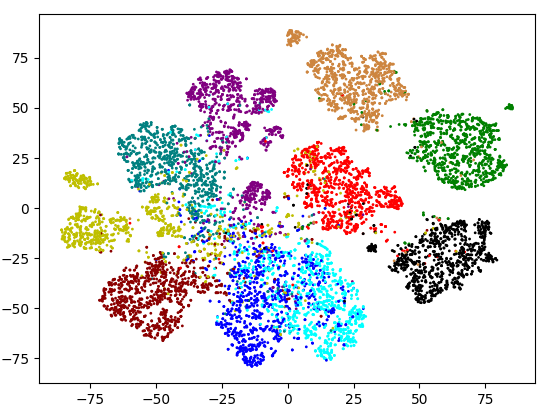}
    \label{fig:b} 
    }

    \subfigure[]{
    \includegraphics[scale=0.29]{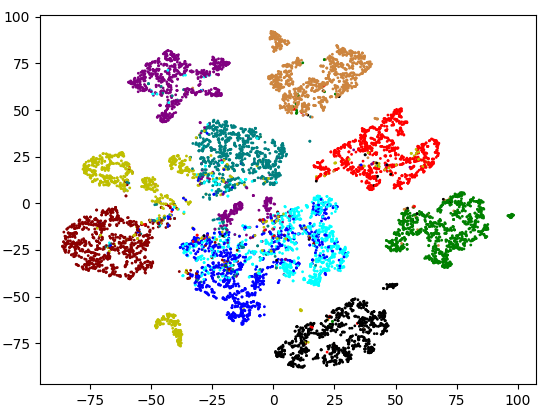}
    \label{fig:c}
    }\hspace{-8pt}
    \subfigure[]{
    \includegraphics[scale=0.29]{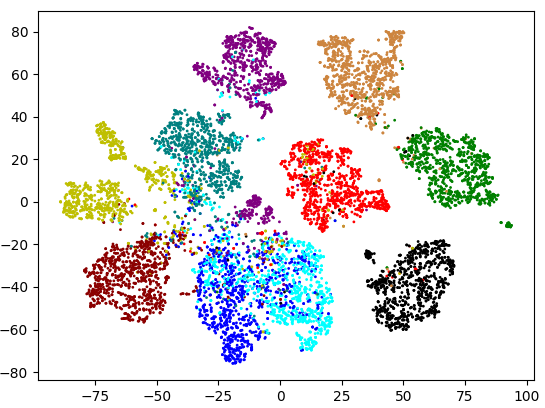}
    \label{fig:d}
    }
    
    \subfigure[]{
    \includegraphics[scale=0.29]{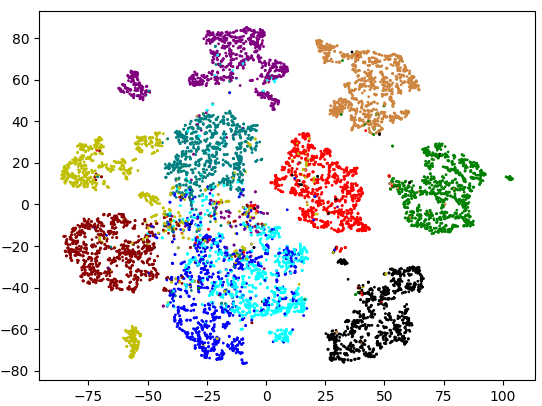}
    \label{fig:e}
    }\hspace{-8pt}
    \subfigure[]{
    \includegraphics[scale=0.29]{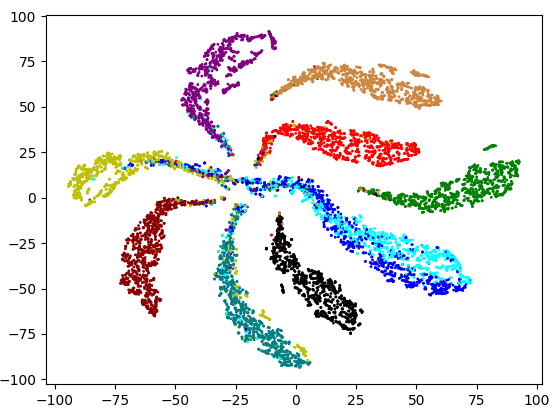}
    \label{fig:f}
    }
\caption{t-SNE visualization of embedding layer and projection layer with different models: (a) is the projection layer of SimCLR; (b) is the embedding layer of SimCLR; (c) is the projection layer of CCL clustering on projection layer; (d) is the embedding layer of CCL clustering on projection layer; (e) is the projection layer of CCL clustering on embedding layer; (f) is the embedding layer of CCL clustering on embedding layer.}
\label{fig:tsne}
\end{figure}
% ----------------------------------------------------

\paragraph{Clustering on Embedding layer vs. Projection layer.}
To verify our assumption about where to cluster, we respectively calculate the cluster-aware contrastive loss using the output of the embedding layer and projection layer. The results are shown in Table \ref{tab:layer}. Apparently, the performance of clustering on the embedding layer is much better than that on the projection layer. 

Moreover, we visualize the features of the two layers through t-SNE to observe the effect of clustering. Figure \ref{fig:a} and \ref{fig:b} show that the feature distributions of SimCLR on two layers are similar. And clustering on the projection layer does not change the feature distribution of the two layers from Figure \ref{fig:c} and \ref{fig:d}. Figure \ref{fig:c} and \ref{fig:a} show because of clustering on projection layer, features on projection layer are denser in the feature space than SimCLR, which makes the performance of OOD detection improve as shown in Table \ref{tab:layer}. Figure \ref{fig:e} and \ref{fig:f} indicate clustering on embedding layer makes feature distribution significantly different on the two layers. From Table \ref{tab:layer} and Figure \ref{fig:tsne}, it can be inferred that clustering on embedding layer causes a clearer feature distribution, which improves the unsupervised OOD detection ability. Overall, clustering and calculating cluster-aware contrastive loss on the embedding layer is better than the projection layer, which is consistent with our assumption.

\paragraph{Update method of cluster centers.}
In Section \ref{sec:method}, we propose CCL needs a warm-up stage and cluster centers need to be updated periodically. The better performance with warm-up in Table \ref{tab:update} indicates only when the backbone encoder is fully capable of instance-level representation can cluster-aware contrastive loss play a stable role, which confirms our viewpoint. Furthermore, we set a series of updating epoch intervals to study the influence of updating frequency. The results show the model has the best performance in unsupervised OOD detection when cluster centers are updated with 10 epoch intervals to calculate the cluster-aware contrastive loss. In practical application, the hyper-parameter needs to be adjusted individually.

%------------------- table----------
\begin{table*}[ht]
\centering
\setlength{\tabcolsep}{6mm}{
    \begin{tabular}{*{6}{c}}
    \toprule
    Number & CIFAR-10 & SVHN & LSUN & ImageNet & CIFAR-100  \\
    \midrule
    & Similarity & \multicolumn{4}{c}{AUROC/Similarity} \\
    \cmidrule(lr){2-2}\cmidrule(lr){3-6}
    {\large 5}     & 0.988  & 98.2 / 0.924  & 96.4 / 0.912  & 93.9 / 0.946   & 91.4 / 0.947   \\
    {\large 10}    & 0.985  & 98.4 / 0.954  & 98.5 / 0.858  & 97.8 / 0.891   & 91.2 / 0.928    \\
    {\large 50}    & 0.771  & 98.4 / 0.588  & 97.4 / 0.662  & 95.4 / 0.713   & 91.2 / 0.718    \\
    {\large 100}   & 0.773  & 98.3 / 0.671  & 98.1 / 0.669  & 96.7 / 0.690   & 91.3 / 0.731    \\
    {\large 500}   & 0.639  & 97.6 / 0.705  & 96.9 / 0.555  & 94.5 / 0.585   & 91.3 / 0.628     \\
    {\large 1000}  & 0.623  & 97.8 / 0.663  & 98.2 / 0.576  & 96.2 / 0.656   & 91.1 / 0.638     \\
    {\large 5000}  & 0.556  & 97.6 / 0.423  & 97.6 / 0.467  & 96.0 / 0.497   & 90.1 / 0.486     \\
    \bottomrule
    \end{tabular}
}

\caption{Comparison of different cluster center numbers. The similarity is the mean of the maximum cosine similarity between embedding layer features and cluster centers.}
\label{tab:cluster}
\end{table*}
%-------------------------------------------------------------------------

\paragraph{The number of cluster centers.} Since we use K-means as clustering method, the number of cluster centers need be determined in advance. We set different numbers of cluster centers to study its influence on CCL. Table \ref{tab:cluster} shows that when it is set to 10, the actual number of CIFAR-10 classes, the model has the best performance. As a result, we think our method could achieve the best performance in practice if there is the prior information about the number of ID sample classes. 

Moreover, to further study the influence of cluster center number on the semantic representation ability of clustering, we calculate the mean of maximum cosine similarity between test samples and cluster centers on the embedding layer. The results in Table \ref{tab:cluster} show that the larger number of cluster centers reduces the similarity with ID samples, which indicates the increase in cluster center number leads to the weakening of semantic representation of cluster centers.

\section{Conclusion}
In this paper, we propose a novel \textit{Cluster-aware Contrastive Learning (CCL)} framework for unsupervised OOD detection. To take advantage of semantic information in comparative learning, we study where and when to cluster to fully extract the latent semantic information between samples. And we design a cluster-aware contrastive loss function to make the model pay attention to the global and local sample relationship. Extensive experiments conducted on seven benchmark datasets demonstrate the effectiveness of our proposed model.

\bibliographystyle{named}
\bibliography{ijcai23}

\begin{thebibliography}{}

\bibitem[\protect\citeauthoryear{Albert \bgroup \em et al.\egroup
  }{2022}]{10-albert2022embedding}
Paul Albert, Eric Arazo, Noel~E O’Connor, and Kevin McGuinness.
\newblock Embedding contrastive unsupervised features to cluster in-and
  out-of-distribution noise in corrupted image datasets.
\newblock In {\em European Conference on Computer Vision}, pages 402--419.
  Springer, 2022.

\bibitem[\protect\citeauthoryear{Bergman and
  Hoshen}{2020}]{goad-bergman2020classification}
Liron Bergman and Yedid Hoshen.
\newblock Classification-based anomaly detection for general data.
\newblock {\em arXiv preprint arXiv:2005.02359}, 2020.

\bibitem[\protect\citeauthoryear{Bergmann \bgroup \em et al.\egroup
  }{2019}]{industrial-bergmann2019mvtec}
Paul Bergmann, Michael Fauser, David Sattlegger, and Carsten Steger.
\newblock Mvtec ad--a comprehensive real-world dataset for unsupervised anomaly
  detection.
\newblock In {\em Proceedings of the IEEE/CVF conference on computer vision and
  pattern recognition}, pages 9592--9600, 2019.

\bibitem[\protect\citeauthoryear{Caron \bgroup \em et al.\egroup }{2020}]{swav}
Mathilde Caron, Ishan Misra, Julien Mairal, Priya Goyal, Piotr Bojanowski, and
  Armand Joulin.
\newblock Unsupervised learning of visual features by contrasting cluster
  assignments.
\newblock {\em Advances in Neural Information Processing Systems},
  33:9912--9924, 2020.

\bibitem[\protect\citeauthoryear{Chan \bgroup \em et al.\egroup
  }{2021}]{autonomous-chan2021entropy}
Robin Chan, Matthias Rottmann, and Hanno Gottschalk.
\newblock Entropy maximization and meta classification for out-of-distribution
  detection in semantic segmentation.
\newblock In {\em Proceedings of the ieee/cvf international conference on
  computer vision}, pages 5128--5137, 2021.

\bibitem[\protect\citeauthoryear{Chen \bgroup \em et al.\egroup
  }{2020}]{simclr}
Ting Chen, Simon Kornblith, Mohammad Norouzi, and Geoffrey Hinton.
\newblock A simple framework for contrastive learning of visual
  representations.
\newblock In {\em International conference on machine learning}, pages
  1597--1607. PMLR, 2020.

\bibitem[\protect\citeauthoryear{Cho \bgroup \em et al.\egroup }{2021}]{mcl}
Hyunsoo Cho, Jinseok Seol, and Sang-goo Lee.
\newblock Masked contrastive learning for anomaly detection.
\newblock {\em arXiv preprint arXiv:2105.08793}, 2021.

\bibitem[\protect\citeauthoryear{Golan and
  El-Yaniv}{2018}]{self2-golan2018deep}
Izhak Golan and Ran El-Yaniv.
\newblock Deep anomaly detection using geometric transformations.
\newblock {\em Advances in neural information processing systems}, 31, 2018.

\bibitem[\protect\citeauthoryear{He \bgroup \em et al.\egroup }{2020}]{moco}
Kaiming He, Haoqi Fan, Yuxin Wu, Saining Xie, and Ross Girshick.
\newblock Momentum contrast for unsupervised visual representation learning.
\newblock In {\em Proceedings of the IEEE/CVF conference on computer vision and
  pattern recognition}, pages 9729--9738, 2020.

\bibitem[\protect\citeauthoryear{Hendrycks \bgroup \em et al.\egroup
  }{2019}]{self1-hendrycks2019using}
Dan Hendrycks, Mantas Mazeika, Saurav Kadavath, and Dawn Song.
\newblock Using self-supervised learning can improve model robustness and
  uncertainty.
\newblock {\em Advances in neural information processing systems}, 32, 2019.

\bibitem[\protect\citeauthoryear{Hsu \bgroup \em et al.\egroup
  }{2020}]{class3-hsu2020generalized}
Yen-Chang Hsu, Yilin Shen, Hongxia Jin, and Zsolt Kira.
\newblock Generalized odin: Detecting out-of-distribution image without
  learning from out-of-distribution data.
\newblock In {\em Proceedings of the IEEE/CVF Conference on Computer Vision and
  Pattern Recognition}, pages 10951--10960, 2020.

\bibitem[\protect\citeauthoryear{Khosla \bgroup \em et al.\egroup
  }{2020}]{supcon}
Prannay Khosla, Piotr Teterwak, Chen Wang, Aaron Sarna, Yonglong Tian, Phillip
  Isola, Aaron Maschinot, Ce~Liu, and Dilip Krishnan.
\newblock Supervised contrastive learning.
\newblock {\em Advances in Neural Information Processing Systems},
  33:18661--18673, 2020.

\bibitem[\protect\citeauthoryear{Lehmann and Ebner}{2021}]{09-lehmann2021layer}
Daniel Lehmann and Marc Ebner.
\newblock Layer-wise activation cluster analysis of cnns to detect
  out-of-distribution samples.
\newblock In {\em International Conference on Artificial Neural Networks},
  pages 214--226. Springer, 2021.

\bibitem[\protect\citeauthoryear{Li \bgroup \em et al.\egroup }{2020}]{pcl}
Junnan Li, Pan Zhou, Caiming Xiong, and Steven~CH Hoi.
\newblock Prototypical contrastive learning of unsupervised representations.
\newblock {\em arXiv preprint arXiv:2005.04966}, 2020.

\bibitem[\protect\citeauthoryear{Li \bgroup \em et al.\egroup
  }{2021}]{21-li2021contrastive}
Yunfan Li, Peng Hu, Zitao Liu, Dezhong Peng, Joey~Tianyi Zhou, and Xi~Peng.
\newblock Contrastive clustering.
\newblock In {\em Proceedings of the AAAI Conference on Artificial
  Intelligence}, pages 8547--8555, 2021.

\bibitem[\protect\citeauthoryear{Liang \bgroup \em et al.\egroup
  }{2017}]{04-liang2017enhancing}
Shiyu Liang, Yixuan Li, and Rayadurgam Srikant.
\newblock Enhancing the reliability of out-of-distribution image detection in
  neural networks.
\newblock {\em arXiv preprint arXiv:1706.02690}, 2017.

\bibitem[\protect\citeauthoryear{Madan \bgroup \em et al.\egroup
  }{2022}]{11-madan2022self}
Neelu Madan, Nicolae-Catalin Ristea, Radu~Tudor Ionescu, Kamal Nasrollahi,
  Fahad~Shahbaz Khan, Thomas~B Moeslund, and Mubarak Shah.
\newblock Self-supervised masked convolutional transformer block for anomaly
  detection.
\newblock {\em arXiv preprint arXiv:2209.12148}, 2022.

\bibitem[\protect\citeauthoryear{Perera \bgroup \em et al.\egroup
  }{2019}]{recons2-perera2019ocgan}
Pramuditha Perera, Ramesh Nallapati, and Bing Xiang.
\newblock Ocgan: One-class novelty detection using gans with constrained latent
  representations.
\newblock In {\em Proceedings of the IEEE/CVF Conference on Computer Vision and
  Pattern Recognition}, pages 2898--2906, 2019.

\bibitem[\protect\citeauthoryear{Reiss and Hoshen}{2021}]{msc-reiss2021mean}
Tal Reiss and Yedid Hoshen.
\newblock Mean-shifted contrastive loss for anomaly detection.
\newblock {\em arXiv preprint arXiv:2106.03844}, 2021.

\bibitem[\protect\citeauthoryear{Ren \bgroup \em et al.\egroup
  }{2019}]{dense2-ren2019likelihood}
Jie Ren, Peter~J Liu, Emily Fertig, Jasper Snoek, Ryan Poplin, Mark Depristo,
  Joshua Dillon, and Balaji Lakshminarayanan.
\newblock Likelihood ratios for out-of-distribution detection.
\newblock {\em Advances in neural information processing systems}, 32, 2019.

\bibitem[\protect\citeauthoryear{Schlegl \bgroup \em et al.\egroup
  }{2017}]{medical-schlegl2017unsupervised}
Thomas Schlegl, Philipp Seeb{\"o}ck, Sebastian~M Waldstein, Ursula
  Schmidt-Erfurth, and Georg Langs.
\newblock Unsupervised anomaly detection with generative adversarial networks
  to guide marker discovery.
\newblock In {\em International conference on information processing in medical
  imaging}, pages 146--157. Springer, 2017.

\bibitem[\protect\citeauthoryear{Sehwag \bgroup \em et al.\egroup
  }{2020}]{23-sehwag2020ssd}
Vikash Sehwag, Mung Chiang, and Prateek Mittal.
\newblock Ssd: A unified framework for self-supervised outlier detection.
\newblock In {\em International Conference on Learning Representations}, 2020.

\bibitem[\protect\citeauthoryear{Sharma \bgroup \em et al.\egroup
  }{2020}]{19-sharma2020clustering}
Vivek Sharma, Makarand Tapaswi, M~Saquib Sarfraz, and Rainer Stiefelhagen.
\newblock Clustering based contrastive learning for improving face
  representations.
\newblock In {\em 2020 15th IEEE International Conference on Automatic Face and
  Gesture Recognition (FG 2020)}, pages 109--116. IEEE, 2020.

\bibitem[\protect\citeauthoryear{Sinhamahapatra \bgroup \em et al.\egroup
  }{2022}]{game}
Poulami Sinhamahapatra, Rajat Koner, Karsten Roscher, and Stephan
  G{\"u}nnemann.
\newblock Is it all a cluster game?--exploring out-of-distribution detection
  based on clustering in the embedding space.
\newblock {\em arXiv preprint arXiv:2203.08549}, 2022.

\bibitem[\protect\citeauthoryear{Tack \bgroup \em et al.\egroup }{2020}]{csi}
Jihoon Tack, Sangwoo Mo, Jongheon Jeong, and Jinwoo Shin.
\newblock Csi: Novelty detection via contrastive learning on distributionally
  shifted instances.
\newblock {\em Advances in neural information processing systems},
  33:11839--11852, 2020.

\bibitem[\protect\citeauthoryear{Wang \bgroup \em et al.\egroup
  }{2022}]{22-wang2022clusterscl}
Yanling Wang, Jing Zhang, Haoyang Li, Yuxiao Dong, Hongzhi Yin, Cuiping Li, and
  Hong Chen.
\newblock Clusterscl: Cluster-aware supervised contrastive learning on graphs.
\newblock In {\em Proceedings of the ACM Web Conference 2022}, pages
  1611--1621, 2022.

\bibitem[\protect\citeauthoryear{Yang \bgroup \em et al.\egroup
  }{2021}]{12-yang2021semantically}
Jingkang Yang, Haoqi Wang, Litong Feng, Xiaopeng Yan, Huabin Zheng, Wayne
  Zhang, and Ziwei Liu.
\newblock Semantically coherent out-of-distribution detection.
\newblock In {\em Proceedings of the IEEE/CVF International Conference on
  Computer Vision}, pages 8301--8309, 2021.

\bibitem[\protect\citeauthoryear{Zhang \bgroup \em et al.\egroup
  }{2021}]{17-zhang2021supporting}
Dejiao Zhang, Feng Nan, Xiaokai Wei, Shangwen Li, Henghui Zhu, Kathleen
  McKeown, Ramesh Nallapati, Andrew Arnold, and Bing Xiang.
\newblock Supporting clustering with contrastive learning.
\newblock {\em arXiv preprint arXiv:2103.12953}, 2021.

\bibitem[\protect\citeauthoryear{Zhong \bgroup \em et al.\egroup
  }{2021}]{16-zhong2021graph}
Huasong Zhong, Jianlong Wu, Chong Chen, Jianqiang Huang, Minghua Deng, Liqiang
  Nie, Zhouchen Lin, and Xian-Sheng Hua.
\newblock Graph contrastive clustering.
\newblock In {\em Proceedings of the IEEE/CVF International Conference on
  Computer Vision}, pages 9224--9233, 2021.

\bibitem[\protect\citeauthoryear{Zhou}{2022}]{20-zhou2022rethinking}
Yibo Zhou.
\newblock Rethinking reconstruction autoencoder-based out-of-distribution
  detection.
\newblock In {\em Proceedings of the IEEE/CVF Conference on Computer Vision and
  Pattern Recognition}, pages 7379--7387, 2022.

\end{thebibliography}

\end{document}